\documentclass[conference]{IEEEtran}
\usepackage{cite}
\usepackage{amsmath,amssymb,amsfonts}
\usepackage{graphicx}
\usepackage{textcomp}
\usepackage{xcolor}
\usepackage{multirow}
\usepackage{float}
\usepackage{acronym}
\usepackage[bb=boondox]{mathalfa}
\usepackage{algorithm} 
\usepackage{algpseudocode} 

\acrodef{DL}[DL]{Deep Learning}
\acrodef{FKV}[FKV]{Facial Kinship Verification}
\acrodef{FIW}[FIW]{Families in the Wild}

\usepackage{changes} 
\definecolor{HB}{RGB}{0,200,0}
\definecolor{NB}{RGB}{200,0,200}
\definechangesauthor[color=HB]{HB}
\definechangesauthor[color=NB]{NB}

\def\BibTeX{{\rm B\kern-.05em{\sc i\kern-.025em b}\kern-.08em
    T\kern-.1667em\lower.7ex\hbox{E}\kern-.125emX}}
\begin{document}

\title{Supervised Contrastive Learning and Feature Fusion for Improved Kinship Verification}

\author{\IEEEauthorblockN{Nazim Bendib}
\IEEEauthorblockA{
\textit{Ecole Nationale Supérieure d'Informatique}\\
Algiers, Algeria \\
jn\_bendib@esi.dz}
}

\maketitle

\begin{abstract}
Facial Kinship Verification is the task of determining the degree of familial relationship between two facial images.
It has recently gained a lot of interest in various applications spanning forensic science, social media, and demographic studies.
In the past decade, deep learning-based approaches have emerged as a promising solution to this problem, achieving state-of-the-art performance.
In this paper, we propose a novel method for solving kinship verification by using supervised contrastive learning, which trains the model to maximize the similarity between related individuals and minimize it between unrelated individuals. Our experiments show state-of-the-art results and achieve 81.1\% accuracy in the Families in the Wild (FIW) dataset.  
\end{abstract}


\section{Introduction}
\ac{FKV} involves comparing the facial features of two individuals and determining whether they are related or not. 
\ac{FKV} is an important research area with diverse applications, such as genealogy research, identifying missing persons, automatically organizing digital photo albums, or as a biometric identifier. 
It requires the ability to identify and analyze the discriminative features that are inherited from parents to their children, or shared between siblings, such as the distance between the eyes, the shape of the nose, and the thickness of the lips. 

Proposed approaches can be grouped into two main categories: 
(1) shallow models \cite{6562692, 7532894, 5652590} that are based on hand-crafted low-level features such as geometric shapes, colors, and distances.
(2) deep models \cite{vuvkou, Hrmann2020AMC, DBLP:journals/corr/abs-2006-00143, DBLP:journals/corr/abs-2006-00154} that leverage the success of deep learning in extracting complex features using Convolutional Neural Networks (CNN). 
This latter generally uses a pre-trained facial recognition model as a baseline to tackle the problem. Indeed, \ac{FKV} is similar to Facial Recognition as both deal with identifying persons based on their facial images. However, it is more challenging to find akin relations due to the potential variance (age, gender, \dots) between individuals of the same family.

Common issues of these conventional approaches are: (1) their susceptibility to variations in facial expressions, lighting, pose, and other factors, which can significantly affect the quality and robustness of the models. (2) most existing works do not take into account the inherent similarities between related individuals, which can lead to suboptimal results. 
While recent research improved on the baseline FKV accuracies, these issues are still a challenge. In this paper, we present a novel natural way of defining FKV using supervised contrastive learning which achieves state-of-the-art results and solves the above-mentioned issues. 
We train a CNN to learn discriminative representations, and map individuals into a representation space where family members are close together. 

The main contributions of this paper are:
\begin{itemize}
    \item To the best of our knowledge, this is the first work to define FKV as a contrastive learning problem with feature fusion. 
    \item We propose a new batch sampling method that does not violate the kinship constraints.
    \item Our methodology is validated on \ac{FIW} benchmark in which we achieve +2\% above state-of-the-art accuracy.
\end{itemize}


\begin{figure*}[t!]
\centering
\includegraphics[width=0.60\textwidth]{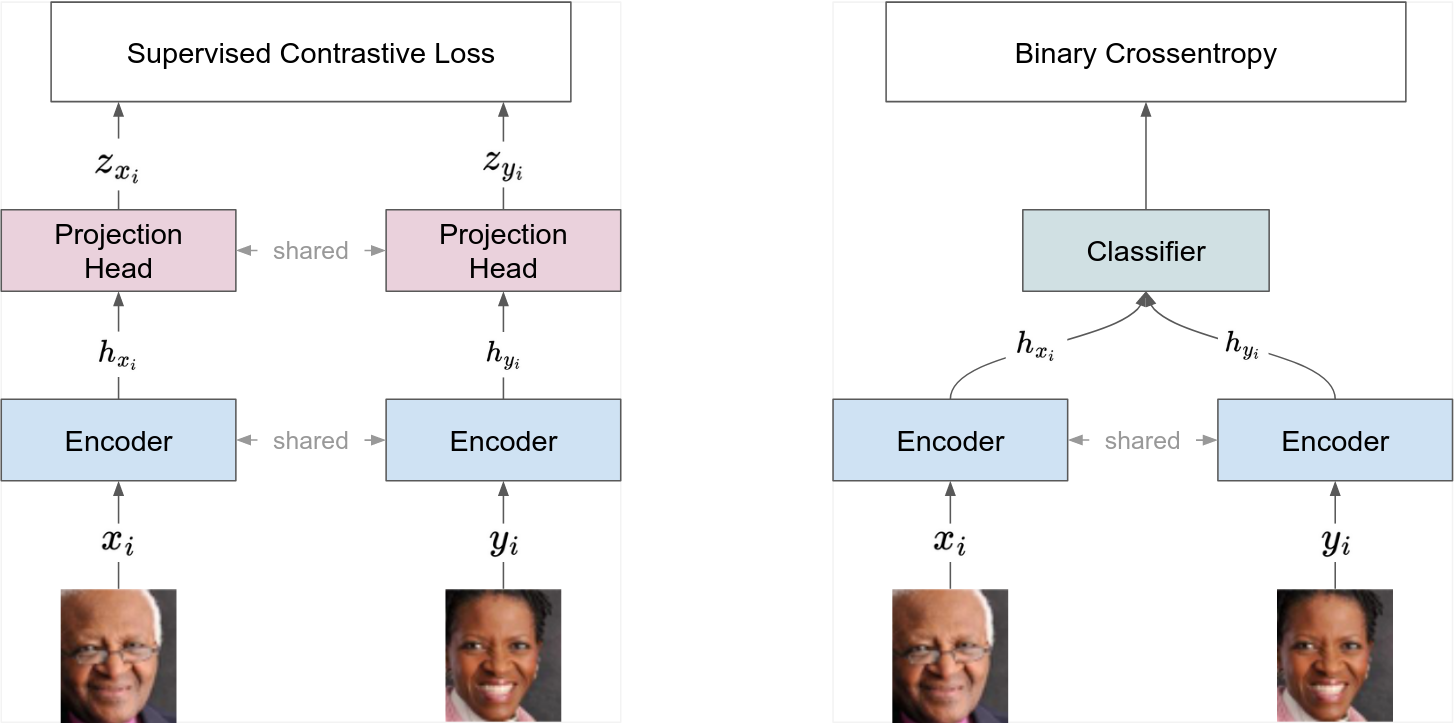}
\caption{First stage (left): the encoder \(f\) is trained by passing the facial representations to the projection head \(g\), which maps them into a 128-dimensional embedding, on which the supervised contrastive loss is applied. Second stage (right): the pre-trained encoder \(f\) is used as a feature extractor for the binary classifier \(d\) .}
\label{fig1}
\end{figure*}

\section{Related Work}
Facial kinship verification has been attracting the attention of researchers since its seminal paper \cite{seminal}. This interest in the problem motivated the development of many labeled family datasets that played a pivotal role in the progress of \ac{FKV}. 

In general, deep learning-based methods achieved much more promising results than handcrafted-based methods, and this is due to using CNNs to exploit large datasets such as FIW. 
A siamese 
network with pre-trained CNNs on facial recognition was a go-to approach for many researchers, even though it naively assumed that the best way to detect kinship is to detect faces that look alike. Other works further improved this by using novel fusion techniques.


Other lines of research in \ac{FKV} focused on using autoencoders, such as \cite{gated} where they implemented a Gated autoencoder to encode faces as genetic features. Other research focused on using GANs \cite{kinshipgan} to synthesize younger versions of an input face to shrink the age gap between samples, leading to a less complex problem.




\begin{figure}[htbp]
\hspace*{-0.5cm}
\centerline{\includegraphics[scale=0.38]{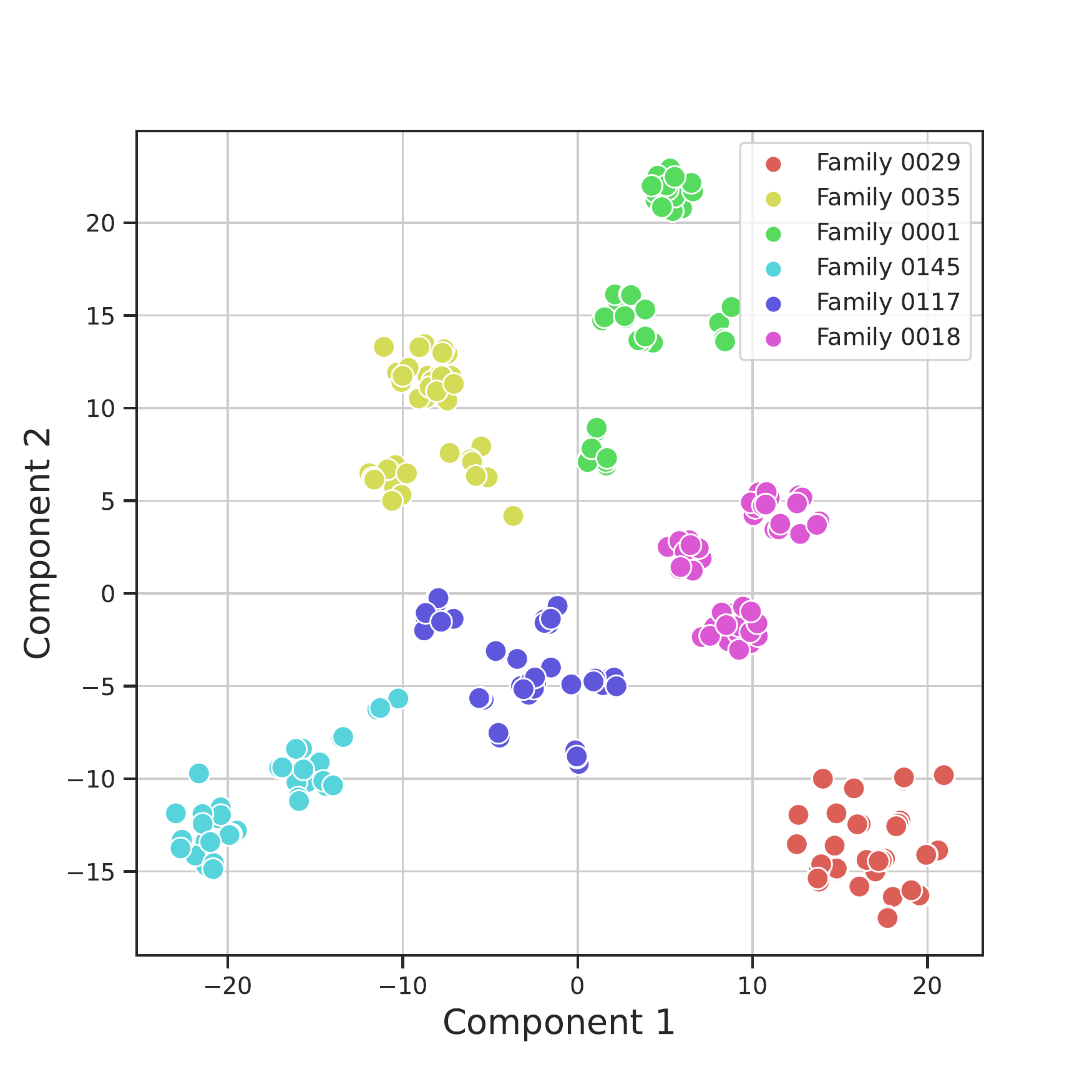}}
\caption{T-SNE visualization of individuals' representations from diverse families.}
\label{fig3}
\end{figure}

\section{Method}
Inspired by the recent success of contrastive learning in computer vision, we propose a novel approach that combines supervised contrastive learning with feature fusion to obtain more discriminative and relevant facial representations. 

We will train our \ac{FKV} model sequentially through two stages, illustrated in figure~\ref{fig1}. 
\begin{itemize}
    \item \textit{Supervised Contrastive Learning:} maximizes the similarity between the representations of related individuals as shown in Fig.~\ref{fig1} (left).
    \item \textit{Binary Classification:} trains a binary classifier on top of the pre-trained feature extractor as shown in Fig.~\ref{fig1} (right), to predict whether two individuals are related.
\end{itemize}

Our framework is composed of the following three components:
\begin{itemize}
    \item \textbf{Neural network encoder \(f(.)\)} that extracts a representation vector from an input facial image.
    \item \textbf{Projection head \(g(.)\)} that maps a representation vector to a lower dimensional space where the contrastive loss is applied.
    \item \textbf{Binary classifier \(d(.)\)} that takes the representations of two input images, applies features fusion, then predicts if they are related or not.
\end{itemize}

In the following, we detail the procedure of each stage.

\begin{table*}[htbp]
\centering
\begin{tabular}{c|cccccccccccc}
\hline
Method                     & BB            & SS            & SIBS            & FS            & FD            & MS            & MD            & GFGD        & GMGD          & GFGS          & GMGS          & Ave.          \\ \hline
Dual-VGGFace \cite{dual}     & 73.0          & 65.8          & 66.9          & 64.0          & 65.2          & 66.2          & 67.4          & /           & /             & /             & /             & 66.9          \\
Benchmark \cite{rob}        & 71.9          & 77.3          & 72            & 68.5          & 69.3          & 69.5          & 71.8          & /           & /             & /             & /             & 71.2          \\
ResNet SDMLoss \cite{cross}   & 72.6          & 79.4          & 74            & 68.0          & 68.3          & 68.8          & 71.3          & /           & /             & /             & /             & 71.2          \\
DeepBlueAI \cite{DBLP:journals/corr/abs-2006-00154}       & 77            & 77            & 77            & 81            & 74            & 74            & 75            & \textbf{79} & 76            & 69            & 67            & 76            \\
Ustc-nelslip \cite{DBLP:journals/corr/abs-2006-00143}     & 75            & 74            & 74            & 82            & 76            & 75            & 75            & \textbf{79} & 76            & 69            & 67            & 76            \\
Vuvko \cite{vuvkou}            & 80            & 80            & 77            & 81            & 75            & 74            & 78            & 78          & 76            & 69            & 60            & 78            \\
Unified Approach \cite{unified} & 85.9          & \textbf{86.3} & 78.0          & 74.9          & 77.4          & 75.6          & 76.9          & /           & /             & /             & /             & 79.3          \\
Ours - Frozen backbone     & 82.2          & 81.4          & \textbf{84.6} & 78.3          & 76.4          & 79.6          & 77.7          & 69.1        & 65.7          & 71.3          & 62.8          & 78.6          \\
Ours - Finetuned backbone  & \textbf{86.2} & 85.1          & 81.6          & \textbf{82.5} & \textbf{78.6} & \textbf{81.9} & \textbf{81.7} & 70.4        & \textbf{76.6} & \textbf{72.9} & \textbf{69.7} & \textbf{81.1} \\ \hline
\end{tabular}
\caption{Detailed and average accuracies of the previous solutions compared to our two models: no finetuning vs with finetuning}
\label{table:resutls}
\end{table*}

\subsection{First stage: Supervised contrastive learning}
Supervised contrastive learning \cite{sup_contrast} is a mixture of contrastive learning~\cite{SSL} and supervised learning. The model is trained on labeled data while using contrastive loss. Our supervised contrastive loss makes the model learn to maximize the similarity between the representations of related individuals and minimize the similarity between unrelated individuals Fig.~\ref{fig3}. 

The encoder \(f(.)\) receives a batch of the form \(\{(x_i, y_i)\}\)  and extracts the facial representations \(\{(h_{x_i},h_{y_i})\}\) where:
\begin{itemize}
    \item \(N\) is the batch size and \(1\le i \le N\).
    \item \((x_i,y_i)\) is the \(i^{th}\) positive pair in the batch.
    \item \(h_{x_i}=f(x_i)\) is the representation of the image \(x_i\).
\end{itemize}
Each representation \(h_k\) is then fed to a projection head \(g(.)\) to map it down into a lower dimensional embedding \(z_k = g(h_k)\). The contrastive loss \cite{simclr} is then applied as follows:
\[ \mathcal{L} = \frac{1}{2N} \sum_{i = 1}^{N} \Big( \mathcal{L}_{scl}(z_{x_i}, z_{y_i}) + \mathcal{L}_{scl}(z_{y_i},z_{x_i}) \Big) \]
where \(\mathcal{L}_{scl}\) is the supervised contrastive loss of a positive pair (related pair):
\[ \mathcal{L}_{scl}(z_i, z_j) = -\log \frac{e^{sim(z_i, z_j)/\tau}}{\sum_{k=1}^{2N}  \mathbb{1}_{[z_i\not=z_k]} e^{sim(z_i, z_k)/\tau} } \]
and \(sim(.,.)\) is a similarity metric. In our approach, we use cosine similarity:
\[  sim(z_i, z_j) = \frac{z_i^Tz_j}{\left\| z_i \right\| \left\| z_j \right\|} \]


\subsection{Second stage: Binary classification}
In this second stage, we train a binary classifier network \(d(.)\) that uses the representations \(h_k\) as input to classify kins and non-kins. These representations are extracted from the pre-trained encoder \(f(.)\) as shown in Fig.~\ref{fig1}(right).
\[ P(kin) = d(h_{x_i}, h_{y_i}) = d(f(x_i), f(y_i)) \]
In order to better capture the underlying structure of the data and extract the most relevant features for the classification, we use feature fusion by calculating multiple quadratic combinations of the representation vectors \cite{DBLP:journals/corr/abs-2006-00154} \( \{(h_i^2-h_j^2), (h_i-h_j)^2\}  \) and then pass them to the classifier. 

\subsection{Batch sampler} \label{sampler}
A key problem in the FIW dataset is that the number of photos per individual and the number of individuals per family are imbalanced, which leads to an exponentially increasing difference in the number of pairs between families. To prevent this, we propose a batch sampling technique that aims to balance the data while assuring that the batch contains distinct families, and no kinship constraints are violated in the process. 
Note that not all individuals in a family are blood-related - A father is non-kin to the mother;  Maternal cousins are non-kin to the father. So unlike 
\cite{vuvkou}
, when we sample a batch of pairs we need to make sure that a pair is a valid kin relationship to not confuse the model with a fake positive pair.

Our batch sampler, see algorithm~\ref{pseudo}, takes a list of pairs, denoted as $relationships$. Each pair contains a list of images and a counter for each image, initially initialized at 0. We first shuffle the list of pairs and then iterate over it to get batches of pairs, i.e., groups of size $batch\_size$. For each batch, we remove any duplicate families, replacing them with other distinct families, 
to not have negative pairs from the same family.
Then, for each pair in the batch, the algorithm selects the image with the minimum counter for each individual and increments its counter. By doing so, it ensures that the batch contains the least-seen images for each family.

\begin{algorithm}
    \caption{Batch Samplers}
    \begin{algorithmic}
        \State \textbf{Input:} relationships, $batch\_size$
        \For{$i \gets 0$ to $len(relationships)$ in batches of size $batch\_size$}
            \State $sub \gets relationships[i:i+batch\_size]$
            \If{not distinct\_families($sub$)}
                \State $sub \gets$ replace\_duplicates($sub$, $relationships$)
            \EndIf
            \State $batch \gets []$
            \For{$pair$ \textbf{in} $sub$}
                \State $per1, per2, fam \gets pair$
                \State $img1 \gets$ get\_image\_with\_min\_count($per1$)
                \State $img2 \gets$ get\_image\_with\_min\_count($per2$)
                \State increment\_count($img1$)
                \State increment\_count($img2$)
                \State $batch$.append(($img1$, $img2$))
            \EndFor
            \State \textbf{yield} $batch$
        \EndFor
    \end{algorithmic}
    \label{pseudo}
\end{algorithm}

\section{Experiments and Results}

In this section, we validate our approach on FIW benchmark. We compare state-of-the-art models to our trained models.
All the work is run on one NVIDIA P100 GPU with 16 GB memory. All the implementation is done using Pytorch.

\subsection{Experiments details}
\paragraph{Architecture}
We use ArcFace100 \cite{arcface} 
as the encoder \(f\), a 2-layer MLP projection head \(g\) that maps facial representations into a 128-dimensional space, and a 2-layer MLP for the binary classifier with sigmoid activation in the last layer. Both the projection head and classifier use ReLU as activation function in the hidden layer. 

\paragraph{Dataset}
Families In the Wild (FIW)~\cite{FIW} is the largest dataset available for facial kinship verification. It contains 11,932 photos of 1,000 families, with at least one photo per individual. This created 656,954 image pairs that can be split over 11 kinship relationships, i.e., father-daughter (FD), father-son (FS), mother-daughter (MD), mother-son (MS), brother-brother (BB), sister-sister (SS), brother-sister (SIBS), grandfather-granddaughter (GFGD), grandfather-grandson (GFGS), grandmother-granddaughter (GMGD), grandmother-grandson (GMGS).

The reason for choosing the FIW dataset was due to two key factors. 
Firstly, The FIW dataset has become widely recognized in the field, establishing itself as a benchmark for evaluating facial recognition algorithms. 
Secondly, the diversity of the data in the FIW dataset makes it a more comprehensive representation of real-world facial images and allows for a more robust evaluation of our approach. This enhances the generalization and reliability of our results and increases their applicability to real-world scenarios.

\paragraph{Preprocessing}
Inspired by the work from \cite{vuvkou}, all the faces are re-detected and aligned using RetinaFace detector to not confuse the backbone with high variance unaligned faces. The images are then normalized and resized to (112,112) to fit the Arcface100 backbone.

\begin{figure}[htbp]
\hspace*{-0.5cm}
\centering
\includegraphics[scale=0.51]{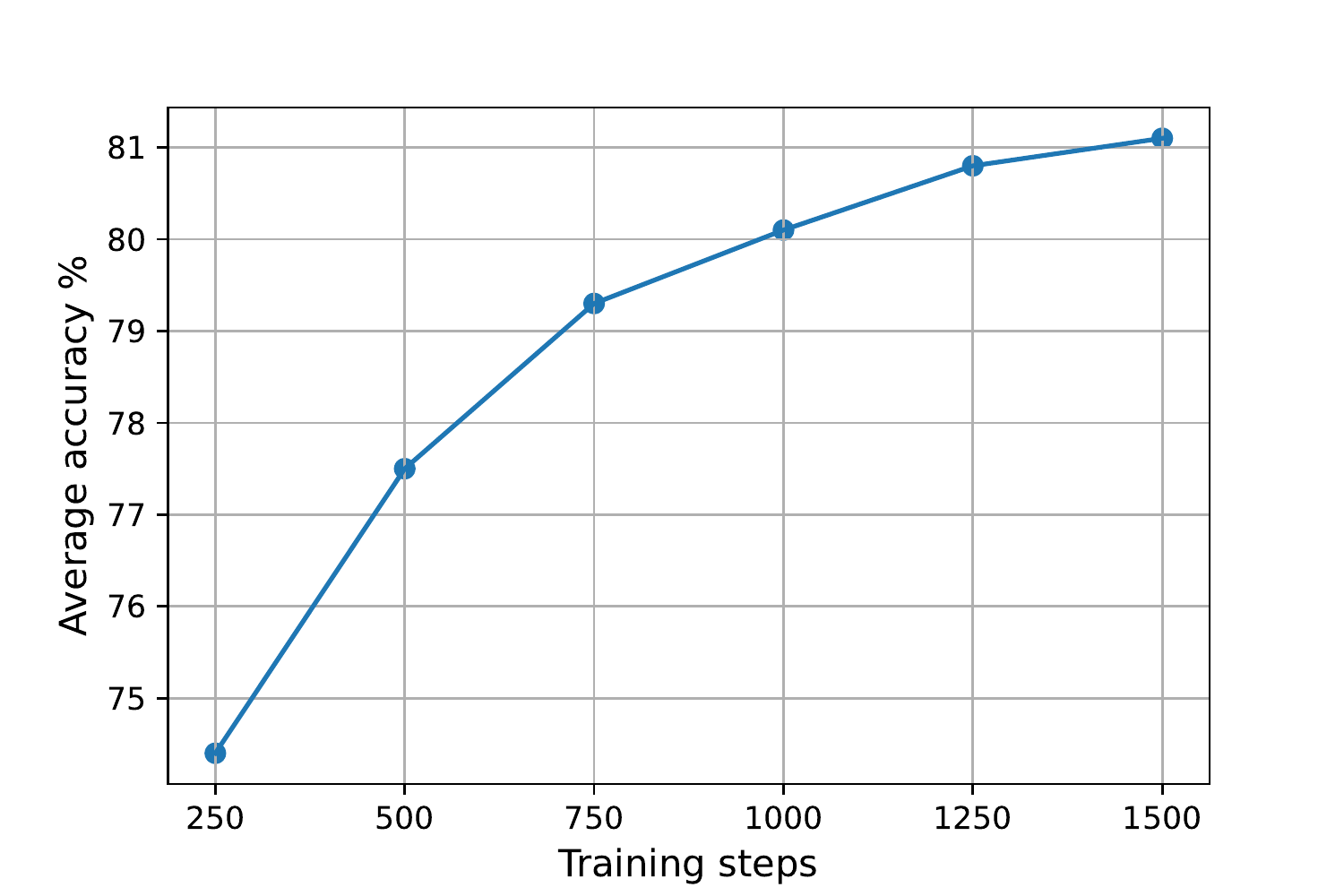}
\caption{Accuracy during the training with supervised contrastive learning fine-tuning}
\label{fig4}
\end{figure}

\paragraph{Hyperparameters}
In the first stage, the model is optimized using SGD with a learning rate of \(5 \times 10^{-5}\), a weight decay of \(10^{-4}\), a momentum of \(0.9\) and a temperature of \(0.07\). In the second stage, the classifier is trained with cross-entropy loss, optimized using Adam with a learning rate of \(10^-4\).

In both stages, we use a batch size of 32 as this is the limit accommodated by our setup.

\paragraph{Data augmentation}
For data augmentation, we use strong color jitter, random grayscale, and random horizontal flipping. This will create more variance in the data and help the model avoid overfitting and generalize better. In contrast to \cite{simclr}, we don't use random cropping, thereby preserving the integrity of the faces and the alignment applied to them.

\subsection{Results}
The experimental results are shown in table~\ref{table:resutls}. Overall, our approach achieves the best performance on multiple kinship types, with a state-of-the-art ~81.1\% average accuracy. 

To evaluate our approach, we tried both freezing and finetuning the pre-trained encoder. Without any fine-tuning and with only 1250 steps of training, our method achieved a near state-of-the-art average accuracy and almost beats the previous best accuracy. This highlights the effectiveness of our feature extraction method in capturing discriminative information.
By finetuning the model, we achieved a new state-of-the-art result in the FIW benchmark, with an average accuracy of 81.1\% while only training for 1500 steps Fig.~\ref{fig4}

Our method has achieved the highest accuracy in almost all relationships. It has demonstrated better performance than previous approaches in \ac{FKV} for GMGD, GFGS, and GMGS, indicating that our method has effectively shrunk the age gap more than the previous ones.

\section{Conclusion}
In this paper, we proposed a novel approach for facial kinship verification based on supervised contrastive learning, a powerful technique that can capture the similarities and differences between related and unrelated pairs of faces. Our approach outperforms other deep learning methods on the Families in The Wild dataset reaching an accuracy of 81.1\%. 
We believe that our training methodology represents a significant step forward in the field of facial kinship verification and demonstrates the power of supervised contrastive learning for capturing complex relationships between faces. 
Looking forward, our work will focus on deploying our contrastive model on edge devices, which can enable real-time facial kinship verification with reduced communication overhead, enhanced privacy and security, and improved user experience.

\bibliographystyle{unsrt}
\bibliography{ref}
\end{document}